\documentclass[11pt,a4paper]{article}
\usepackage{authblk}

\usepackage{enumerate}
\usepackage{graphicx}
\usepackage{mathtools}
\usepackage{graphicx}
\usepackage[hyperref]{eacl2021}
\usepackage{times}
\usepackage{latexsym}

\usepackage{microtype}

\aclfinalcopy % Uncomment this line for the final submission
\title{Persian-WSD-Corpus: A Sense Annotated Corpus for Persian All-words Word Sense Disambiguation}

\author[1]{Hossein Rouhizadeh}
\author[1]{Mehrnoush Shamsfard }
\author[2]{Vahideh Tajalli}
\author[3]{Masoud Rouhziadeh}

\affil[1]{Shahid Beheshti University, Tehran, Iran}
\affil[2]{University of Tehran, Tehran, Iran}
\affil[3]{Johns Hopkis University, MD, USA}
\affil[ ]{ {hrouhizadeh@gmail.com, m-shams@sbu.ac.ir}}
\affil[ ]{ {vtajalli@gmail.com, mrouhizadeh@gmail.com}}

\date{}

\begin{document}
\maketitle
\begin{abstract}
Word Sense Disambiguation (WSD) is a long-standing task in Natural Language Processing(NLP) that aims to automatically identify the most relevant meaning of the words in a given context. Developing standard WSD test collections can be mentioned as an important prerequisite for developing and evaluating different WSD systems in the language of interest. Although many WSD test collections have been developed for a variety of languages, no standard All-words WSD benchmark is available for Persian. In this paper, we address this shortage for the Persian language by introducing SBU-WSD-Corpus, as the first standard test set for the Persian All-words WSD task. SBU-WSD-Corpus is manually annotated with senses from the Persian WordNet (FarsNet) sense inventory. To this end, three annotators used SAMP (a tool for sense annotation based on FarsNet lexical graph) to perform the annotation task. SBU-WSD-Corpus consists of 19 Persian documents in different domains such as Sports, Science, Arts, etc. It includes $5892$ content words of Persian running text and $3371$ manually sense annotated words ($2073$ nouns, $566$ verbs, $610$ adjectives, and $122$ adverbs). Providing baselines for future studies on the Persian All-words WSD task, we evaluate several WSD models on SBU-WSD-Corpus. The corpus is publicly available at https://github.com/hrouhizadeh/SBU-WSD-Corpus.

\end{abstract}
\section{Introduction} \label{sec:intro}
Word Sense Disambiguation (WSD) is an open problem in Natural Language Processing (NLP) which aims to automatically recognize the correct meaning of ambiguous words in a particular context. For instance, consider the sentence 
"$The\ bank\ will\ lend\ us\ money.$", where we want to disambiguate the word $bank$. Retrieving all possible meanings of $bank$ from a pre-defined sense inventory (WordNet, for instance), a WSD algorithm should be ideally able to associate the word $bank$ with its $financial\ institute$ meaning.\newline WSD has applications in other NLP tasks such as Machine Translation \citep{carpuat2007improving}, Information Retrieval and Extraction \citep{zhong2012word}, Question Answering \citep{ramakrishnan2003question}, etc. WSD tasks can be distinguished into two generic categories: (1) Lexical Sample WSD and (2) All-words WSD. Developed Lexical Sample WSD systems aim to disambiguate a set of restricted predefined words. Whereas the goal of developing All-words WSD systems is disambiguating all occurring words in a particular context. Generally, All-words WSD approaches are useful for downstream NLP applications \citep{saeed2019sense}. Compared to the Lexical Sample approaches, developing such systems seems to be more challenging. This is mainly because the developed All-words WSD systems should ideally be able to cover a wide range of open-class words in the language of interest. Whereas, the Lexical Sample systems only require disambiguating a limited number of words. In this paper, we focus on All-Words WSD for the Persian language. \newline
WSD approaches can be grouped into two main approaches: (1): knowledge-based and (2): supervised. Knowledge-based WSD approaches exploit information from a lexical resource such as machine-readable dictionaries, thesauri, and ontology to perform WSD. On the other hand, supervised systems apply machine learning techniques on a sense-annotated corpus to train WSD models. Thanks to the training phase, supervised systems generally outperform knowledge-based alternatives. It worth noting that, due to the unavailability of sense-annotated corpora for many languages, performing supervised WSD is not possible. On the other hand, knowledge-based approaches only require lexical resources that are available for a wide range of languages and can be used as an appropriate alternative. To the best of our knowledge, the only available sense-annotated corpus for the Persian language is Persian SemCor \citep{rouhizadeh-etal-2021-persian}, which have been developed automatically.
Previous studies on All-words WSD have focused on a variety of languages such as English, Dutch, Italian, etc \citep{oele2017distributional,popov2019know,raganato2017neural}. However, many low-resource languages such as Persian have not been studied as well. In this paper, we introduce and discuss the creation pipeline of SBU-WSD-Corpus as the first developed test set for the Persian All-words WSD.  \newline
Persian (also known as Farsi) is an Indo-European (IE) language that is currently spoken by more than $110$ million people in several countries such as Iran, Afghanistan, and Tajikistan. Persian language uses a modified Arabic script and is written from right to left. Millions of Persian texts are available via online web pages, newspapers, books, etc. As a result, there is no doubt in the necessity of developing computational models for Persian as a low-resource language\citep{shamsfard2011challenges}. Similar to other fields of study, standard test sets are required for evaluating WSD approaches. However, none is available for Persian. The main objective of this research is to address the lack of an All-words WSD test set for the Persian language.\newline
SBU-WSD-Corpus contains $5892$ content words of Persian running text. The corpus includes $3371$ instances ($2073$ nouns, $566$ verbs, $610$ adjectives, and $122$ adverbs) which are manually annotated by three annotators. We benchmark SBU-WSD-Corpus with several supervised and knowledge-based WSD models, providing baseline results for future research on All-Words WSD for the Persian language.\newline
The main contributions of this research are as follows  :
\begin{enumerate}
    \item \textbf{Creating a standard All-words WSD data set} \newline
    With the goal of developing an standard All-Words WSD data set, we followed all guidelines, suggested by SensEval-2 \citep{edmonds2001senseval}. To the best of our knowledge, this is the first available test set for Persian All-words WSD task. With the introduction of SBU-WSD-Corpus, we hope to open avenues for future WSD research in Persian. Additionally, we provide details of our corpus creation pipeline, which can be useful for researchers of other low resource languages to develop similar useful resources.    
    \item \textbf{Presenting benchmarks for future research in Persian All-Words WSD}\newline
    To provide baseline for evaluation of Persian All-Words WSD systems, a set of best performing supervised (trained on Persain SemCor) and knowledge-based WSD systems are carried out on SBU-WSD-Corpus. In addition, detailed analysis and comparison between different systems are provided.
    \item \textbf{Usefulness of SBU-WSD-Corpus for evaluation of other Persian NLP tasks}\newline
    The whole documents of SBU-WSD-Corpus has been manually tokenized, Pos-tagged and lemmatized by an expert linguist. As a result, it can be also used as a test set for evaluating a range of basic Persian preprocessing tools such as PoS-taggers, lemmatizers, tokenizers and sentence segmentations, etc. 
    \item \textbf{Free access to the  developed data set}\newline 
    To encourage future research on Persian All-words WSD, SBU-WSD-Corpus will be freely available for the research community. 
\end{enumerate}

The rest of the paper is structured as follows. Section $2$ surveys a range of related works. Section $3$ describes the different steps of creating the corpus. Section $4$ introduces the WSD experiments applied to the corpus. In section $5$ the results and the analysis about the performance of the evaluated benchmarks are presented. Finally, the conclusions and further possible works are found in section $6$. 

\section{Related Work}

Over recent decades, a variety of sense annotated 
corpora have been developed for both All-words and Lexical-Sample WSD tasks. Generally, sense annotated corpora can be divided into two main groups: (a)\textbf{WSD Training corpora} and (b)\textbf{WSD Test Set corpora}. 
\begin{itemize}
% http://www.wsdbook.org/resources.html
  \item  \textbf{WSD Training corpora}, which includes a variety of sense annotated samples in the language of interest. Sense annotated corpora for lexical sample task only include annotated samples for the limited number of predefined words. However, All-words sense annotated corpora should ideally cover multiple instances for a wide range of open-class words. \newline
    Among the developed training WSD datasets, we briefly introduce SemCor \citep{miller1994using} and its different versions, OMSTI \citep{taghipour2015one}, 
    the Italian Syntactic-Semantic Treebank \citep{montemagni2003building}, CLE Urdu Sense Tagged corpus \citep{urooj2014sense} as All-words WSD datasets and DSO corpus \citep{ng1999case} , Line-hard-Serve corpus \citep{miller1993semantic} and the Interest corpus \citep{bruce1994word} as Lexical Sample WSD datasets in the following. \newline
    SemCor, is the first and most prominent All-words Sense annotated corpora for English. SemCor contains $352$ manually tagged documents (Taken from Brown corpus \citep{francis1979brown}) and includes $226040$ sense annotations. It was initially tagged with senses from WordNet $2.1$. Sense tags of the current version of SemCor, are mapped to WordNet $3.0$ senses. Different versions of SemCor are also available for some other languages. Jsemcor \citep{bond2012japanese}, Eusemcor \citep{agirre2005eusemcor}, Bsemcor \citep{koeva2010bulgarian} , and Spsemcor \citep{izquierdo2006spanish} are developed versions of Semcor for Japanese, Basque, Bulgarian and Spanish languages respectively. OMSTI (One Million Sense-Tagged Instances) is another widely used All-words sense annotated corpora for English. It was semi-automatically annotated with senses from WordNet $3.0$ and includes $911134$ sense annotations in $813798$ sentences. An English-Chinese parallel corpus \citep{eisele2010multiun} is used for the construction of OMSTI.
    The Italian Syntactic-Semantic Treebank (ISST) is an Italian manually All-words sense-annotated corpus. The corpus consists of $305547$ tokens including $81236$ manually sense tagged words, annotated with Italian WordNet \citep{roventini2003italwordnet}. CLE Urdu Digest corpus is the Urdu All-words WSD corpus which contains $17006$ sense annotated nouns, tagged with senses from CLE Urdu WordNet \citep{urooj2014sense}.\newline
    The Lexical sample sense annotated corpora surveyed in this section include DSO, Line-hard-Serve and the Interest corpus.
    DSO corpus is a manually sense-annotated corpus including $192800$ sentences drawn from Brown corpus and  Wall Street Journal. $121$ nouns and $70$ verbs have been tagged with senses from WordNet $1.5$. Line-hard-Serve is another predominant English lexical-sample corpus. It includes $12000$ instances from the American Printing House for the Blind, and the San Jose Mercury of the words $line$ (noun), $hard$ (adjective), and $serve$ (verb).
        \item \textbf{WSD Test Set corpora}, which are not as large as training corpora and as a result, are not appropriate for use as training sets in supervised approaches.\newline  
 The major part of developed WSD benchmark corpora for both Lexical Sample and an All-words WSD tasks belongs to SensEval (The International Workshop on Evaluating Word Sense Disambiguation Systems) and SemEval (The International Workshop on Semantic Evaluation) competitions. SemEval (the new name of Senseval) is an ongoing series of evaluations of computational semantics systems in several languages. The main focus of Senseval-1 through Senseval-3 was on both All-words and Lexical Sample WSD tasks. The fourth series of Senseval (renamed Semeval) has been expanded to the evaluation of computational semantic analysis systems not necessarily related to WSD. The outcomes of the competitions are standard WSD frameworks for multiple languages \citep{edmonds2001senseval,navigli2013semeval,moro2015semeval}. For instance, the main benchmark for English All-words WSD (presented by \cite{raganato2017word}) is the unified version of different Senseval and Semeval English All-words WSD tasks \citep{edmonds2001senseval,snyder2004english,pradhan2007semeval,navigli2013semeval,moro2015semeval}. It contains $7253$ sense-annotated instances ($4200$ nouns, $1652$ verbs, $955$ adjectives and $346$ adverbs) annotated with senses from WordNet $3.0$ sense inventory. A variety of European languages such as English, French, Dutch, Italian and Spanish are covered by different series of theses competitions.\newline 
 Recently, \cite{saeed2019word} and \cite{saeed2019sense} developed a Lexical Sample and All-words test sets for the Urdu language. The Lexical Sample corpus includes sense annotated samples for $50$ target words ($30$ nouns, $11$ adjectives, and $9$ adverbs) and the All-words corpus contains $5042$ words of Urdu running text and $466$ sense annotated words. Ambiguous words within both corpora were manually tagged with senses from the Urdu Lughat dictionary. 
\end{itemize}
Although several WSD training and test corpora have been developed for a variety of languages, no WSD corpus is available for Persian. To address the lack of a standard WSD benchmark for Persian, we put forward SBU-WSD Corpus as the first Persian All-words WSD test set. We also carried out a set of best performing WSD systems on SBU-WSD-Corpus as baseline for future researches in Persian All-words WSD. 
\begin{table*}[]
\centering
\begin{tabular}{|c|c|c|c|}
\hline
\begin{tabular}[c]{@{}c@{}}Relation \\ name\end{tabular} & \begin{tabular}[c]{@{}c@{}}Relation \\ Type\end{tabular} & Definition                                                                               & \begin{tabular}[c]{@{}c@{}}Example\\  in WorNet\end{tabular}                            \\ \hline
Hypernymy                                                & Semantic                                                 & \begin{tabular}[c]{@{}c@{}}A is hypernym of B,\\ if every B is a kind of A\end{tabular}  & \begin{tabular}[c]{@{}c@{}}A Portable Computer \\ is Hypernym of \\ Laptop\end{tabular} \\ \hline
Hyponymy                                                 & Semantic                                                 & \begin{tabular}[c]{@{}c@{}}A is hyponym of B,\\ if every A is a kind of B\end{tabular}   & \begin{tabular}[c]{@{}c@{}}Olive \\ is Hyponym of \\ Fruit\end{tabular}                 \\ \hline
Antonymy                                                 & Lexical                                                  & \begin{tabular}[c]{@{}c@{}}A is Antonym of B,\\ if A is polar opposite of B\end{tabular} & \begin{tabular}[c]{@{}c@{}}Hot \\ is Antonym of\\  Cold\end{tabular}                    \\ \hline
\end{tabular}
\caption{Example of some WordNet relations.}
\label{tab:wn_rels}
\end{table*}

\section{Building the SBU-WSD-Corpus}
To create a standard All-words WSD test set, we followed the suggestions made by the Senseval-2 \citep{edmonds2001senseval} competition. For the All-words task, the Senseval-2 guidelines suggest that (1) a standard test set should contain at least $5000$ words of the running text, and (2) all context words should be tagged. The creation of SBU-WSD-Corpus can be thought of as a pipeline of four steps(i.e, Data Collection , Choosing sense inventory, Annotation process and Corpus Format), 
described in the following sections (Sections \ref{sec:da-sel} to \ref{sec:cor_format}).The statistics of SBU-WSD-Corpus are presented in Section \ref{sec:cor_sta}.
\subsection{Data collection} \label{sec:da-sel}
The documents selected for the SBU-WSD-Corpus are taken from in-house news corpora which include one million news documents crawled from different Iranian news websites. The news corpora contain documents from a variety of domains including sports, political, science, cultural, etc.\newline
The process of collecting documents for SBU-WSD-Corpus includes two steps:\newline 
We first extracted $100$ documents from our news corpora and then computed the average ambiguity of the context words of each document. Second, in order to make the task more challenging, we chose $19$ documents with highest average ambiguity for construction of SBU-WSD-Corpus. As prepsocessing step, we first manually tokenized, lemmatized, and PoS-tagged the documents to make them ready for the sense annotation phase. 
\subsection{Sense Inventory}\label{sec:SI}
WordNet \citep{miller1990introduction} is one of the most widely used lexical resources in many areas of NLP including WSD. It was originally designed for English at Princeton University. The basic components of WordNet are synsets, each expressing a unique concept by a set of words with the same meaning and PoS, a gloss (i.e, a brief definition of the synset words), and possibly an example (i.e, a usage example of synsets words). WordNet entries are represented by different synsets, denoting the different meanings they can take. For instance
%as shown in Table\ref{syns_of_phone}, 
the word $phone$ has four synsets in WordNet, denoting four possible meanings of $phone$ in multiple contexts\footnote{In table, each synset are shown in the W\#N\#i or W\#V\#i format which correspond to the ith nominal or verbal synsets of the target word W in the WordNet, respectively,}. The current version of WordNet (WordNet 3.1) covers $155,327$ English words and phrases organized in $117,979$ synsets. \newline
WordNet synsets are interlinked via Lexical or Semantic relations which are held between pairs of word senses and synsets, respectively. WordNet can also be viewed as a semantic network in which nodes correspond to the synsets and edges to the lexical or semantic relations. Instances of Lexical and Semantic relations are shown in table \ref{tab:wn_rels}. \newline

\begin{figure*}
\begin{center}
  \includegraphics[width=0.99\textwidth]{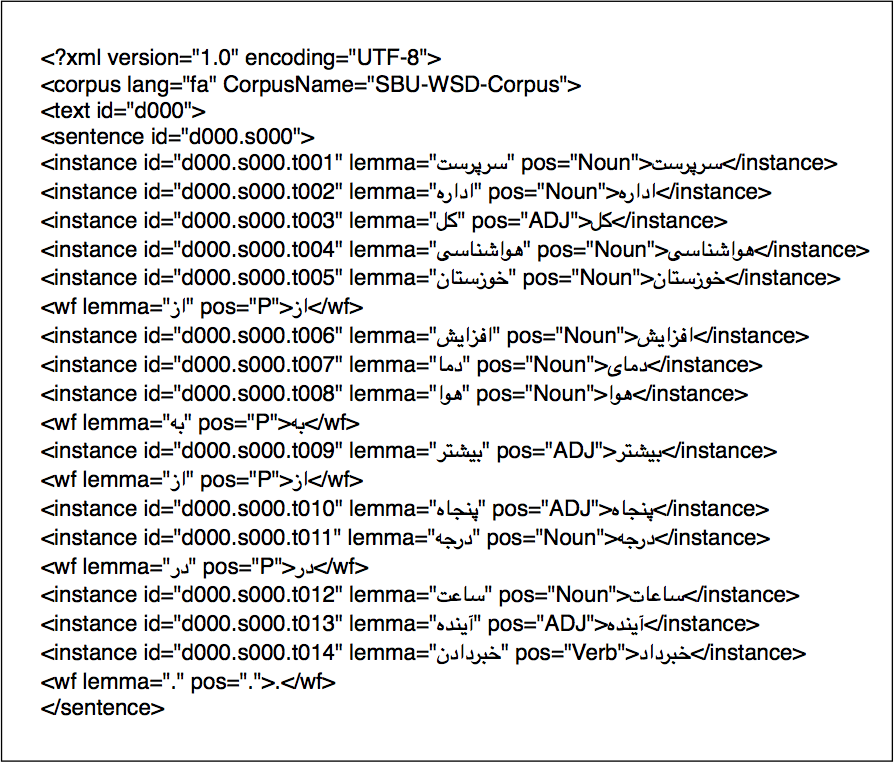}
  \caption{An example sentence drawn from SBU-WSD-Corpus}     
  \label{sample}
  \end{center}
\end{figure*}

\textbf{FarsNet: The Persian WordNet}\newline
Currently, WordNet is developed for many languages including Persian. The Persian WordNet, FarsNet \citep{shamsfard2010semi}, is the first lexical ontology for the Persian language which has been developed in the NLP lab of Shahid Beheshti University. The FarsNet project developed for more than 12 years. Over two past decades, a range of development have been done on FarsNet \citep{rouhizadeh2007building, rouhizadeh2010developing, khalghani2018extraction}. The current version of FarsNet (FarsNet $3.0$) covers more than $100,000$ Persian words and phrases and $40,000$ synsets \footnote {FarsNet web service is freely available at farsnet.nlp.sbu.ac.ir}. Similar to other WordNets, FarsNet groups words (nouns, verbs, adjectives, and adverbs) into synsets and connect them via different kinds of relations. FarsNet also provides a gloss and an example for each synset.
\begin{table*}
\centering
\begin{tabular}{|c|c|c|c|c|} 
\hline
\multicolumn{1}{|l|}{}          &            & Test Set & Tuning Set & All   \\ 
\hline
\# Docs                         &            & 13       & 3          & 16    \\ 
\hline
\# Tokens                       &            & 5045     & 847        & 5892  \\ 
\hline
\multicolumn{1}{|l|}{Number of} & Nouns      & 1764     & 307        & 2071  \\ 
\cline{2-5}
Instances                       & Verbs      & 494      & 70         & 564   \\ 
\cline{2-5}
per                             & Adjectives & 515      & 95         & 610   \\ 
\cline{2-5}
PoS                             & Adverbs    & 111      & 11         & 122   \\ 
\hline
Mean                            & Nouns      & 4.0      & 3.9        & 4.0   \\ 
\cline{2-5}
Sense                           & Verbs      & 3.4      & 2.9        & 3.3   \\ 
\cline{2-5}
per                             & Adjectives & 1.6      & 1.7        & 1.6   \\ 
\cline{2-5}
PoS                             & Adverbs    & 1.2      & 1.3        & 1.2   \\
\hline 
\end{tabular}
\caption{General statistics of SBU-WSD-Corpus}
\label{general_sta_table}
\end{table*}
FarsNet relations can be classified into two major groups: inner-language and inter-language relations.\newline
The inner-language relations are defined between FarsNet senses and synsets while the inter-language relations align the FarsNet and WordNet synsets. The inner language relations of FarsNet include all WordNet $2.1$ relations (i.e. hypernymy, hyponymy, holonymy, antonymy, etc.) as well as some extra relations such as agent-of, patient-of, salient, etc. \newline 
Additionally, as FarsNet $3.0$ is mapped to WordNet $3.0$, 
the inter-language relations (equal-to and near-equal- to) are defined between FarsNet and WordNet $3.0$ synsets. \newline
In this research, we used FarsNet as the sense inventory to annotate the context words of the documents.

\subsection{Annotation Process} \label{sce:anot_pro}
The whole SBU-WSD-Corpus is manually annotated by three Persian native speakers. All the annotators were familiar with FarsNet and WSD. To achieve a high-quality sense-annotated corpus, we followed the annotation procedure, suggested by \cite{saeed2019sense}. The annotation process consists of two steps. In the first step, two taggers used SAMP (a tool for sense annotation with senses from FarsNet $3.0$ \footnote{The tool is available at: http://nlplab.sbu.ac.ir:2347/tagger/} ) to annotate $3$ documents of the corpus. An expert linguist together with both annotators then discussed the annotations specifically the conflicting ones. Taggers then annotated the rested documents.\newline
As the final phase, the expert linguist checked all annotations and re-annotated the words with different sense labels. The Inter-Annotator Agreement (IAA) and Cohen's kappa score obtained from the first step were 90.3 and 0.83 respectively.

\subsection{Corpus Format} \label{sec:cor_format}
The corpus is released in a standard XML format (from \cite{raganato2017word}) including a single file in which all the documents are stored in. A part of the corpus is shown in Figure \ref{sample}. In the following, we describe the XML tags of the corpus. 
\begin{itemize}
    \item $<$corpus$>$: The tag indicates the beginning of the whole corpus. 
 \item $<$text id$>$: The $<$text id$>$ tag is representative of the beginning of a new document each specified with a unique identifier attribute (id). 
\item  $<$sentence$>$: similar to $<$text id$>$, $<$sentence$>$ tag shows the start of a particular sentence specified with a unique id attribute. 
\item $<instance>$: The tag represents context words with a relevant sense in FarsNet and specifies unique id, lemmas (Lemma), and PoS tags. 
\item $<$wf$>$: $<$wf$>$ tag shows a context word with no corresponding sense in FarsNet, specified with a lemma (Lemma) and PoS tag. 
\end{itemize}
\subsection{Corpus Statistics} \label{sec:cor_sta}
\begin{figure}
\begin{center}
  \includegraphics[width=0.5\textwidth]{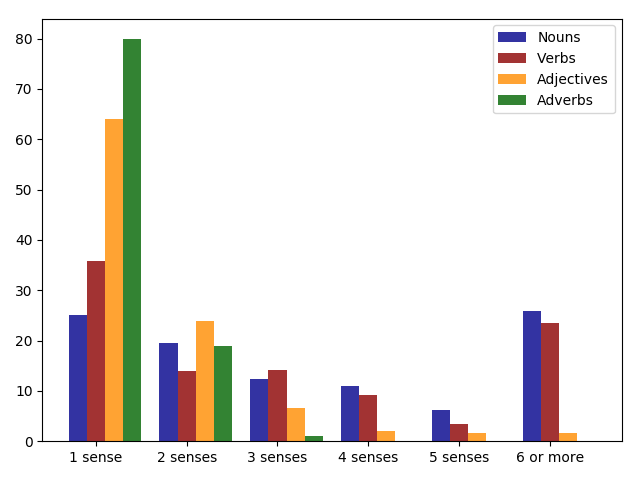}
  \caption{Sense distribution of the annotated words of the SBU-WSD-Corpus, divided by PoS tags.}     
  \label{birds}
  \end{center}
\end{figure}

SBU-WSD-Corpus consists of $19$ documents obtained from in-house news corpora. The documents cover different domains including sports, religion, culture.  In table \ref{general_sta_table} we show the general statistics of the dataset. For both test and tuning set, we report the number of words of running text together with the number of annotated words and ambiguity level per PoS. Following WSD literature, we computed ambiguity level of as total number sense candidates of words, divided by the number of annotated words. It worth noting that monosemous instances have been considered in the process. We also show the sense distribution of the test set words per PoS in Figure \ref{birds}. As it can bee seen, nouns are the most ambiguous part-of-speech followed by verbs, adjectives and adverbs which shows the least ambiguity in their meaning. In addition, more than $25$ percent of nouns and $23$ percent of verbs have more than $5$ different meanings in FarsNet, indicating the task hardness on disambiguating nouns and verbs of the developed corpus. On the other hand, adjectives and adverbs seem easier to disambiguate, as most of them have only one or two senses in FarsNet.\newline
\def\sc{.7}
\def\sc{.7}
% \begin{figure}
% \begin{center}
%   \includegraphics[width=0.5\textwidth]{plot.jpg}
%   \caption{Sense distribution of the annotated words of the SBU-WSD-Corpus, divided by PoS tags.}     
%   \label{birds}
%   \end{center}
% \end{figure}
\section{Experimental Setup}
In this section, we present several supervised and knowledge-based systems as baselines of Persian All-words WSD task. The systems are introduced in section \ref{sec:eval_sys}, the evaluation measures are explained in section \ref{eval_mes} and the results and analysis about the performance of the systems are shown in section \ref{sec:res}.

\subsection{Comparison Systems} \label{sec:eval_sys}
In this section, we briefly describe the All-words WSD systems used in our experiments. We include 10 systems (five supervised and five knowledge-based) in our empirical comparison. 

% \bf{Knowledge-based Systems}
\begin{itemize}
    \item FarsNet 1st sense \newline
    As mentioned in \cite{shamsfard2010semi}, FarsNet word senses are ranked by their use in Persian by expert linguists. We consider the FarsNet 1st sense approach as the baseline of knowledge-based systems. The approach is context-independent and always chooses the first sense of FarsNet as the most probable sense of each context word. 
    \item  Lesk and Extended Lesk \newline
Lesk (Lesk, 1986) is one of the most traditional WSD algorithms based on the overlap between the definition of senses and the context words. The algorithm counts the mutual words between the gloss a given sense and the context of the target word and chooses the sense with the highest count as the proper one. An extention of the Lesk algorithm has been developed by Banerjee and Pedersen (2003). The pipeline of the proposed algorithm (named as Extended Lesk) is highly similar to the Lesk algorithm. The only difference is that Extended Lesk expands the definition of a given sense by including the definitions of its semantically related concepts from WordNet (e.g. hypernyms, hyponyms, etc). 
\begin{table*}
\centering
\begin{tabular}{|c|c|c|c|c|c|c|} 
\hline
Approach   & System        & Noun & Verb & Adjective & Adverb & All   \\ 
\hline
           & MFS           & 59.2 & 65.0 & 84.2      & 90.1   & 65.8  \\ 
\cline{2-7}
Supervised & MLP           & 64.9 & 73.1 & 89.5      & 90.1   & 72.4  \\ 
\cline{2-3}\cline{5-7}
Sytems     & DT            & 63.2 & 71.5 & 90.1      & 90.1   & 70.6  \\ 
\cline{2-7}
           & KNN           & 64.8 & 73.7 & 90.2      & 90.1   & 71.4  \\ 
\hline
           & SVM           & 65.0 & 65.0 & 90.0      & 90.1   & 72.7  \\ 
\cline{2-7}
Knowledge  & FN 1st Sense  & 48.4 & 43.5 & 81.1      & 90.0   & 55.0  \\ 
\cline{2-7}
Based      & Basile14      & 62.7 & 66.3 & 83.6      & 82.9   & 67.8  \\ 
\cline{2-7}
Systems    & UKB (ppr)     & 58.4 & 70.5 & 82.4      & 83.6   & 65.7  \\ 
\cline{2-7}
           & UKB (ppr-w2w) & 58.3 & 71.5 & 84.4      & 84.5   & 66.2  \\
\hline
\end{tabular}
\caption{F-1 performance of different supervised and knowledge-based models on SBU-WSD-Corpus}
\label{table_result_111}
\end{table*}
    \item Basile14 \newline
    Extending two aforementioned variations of Lesk algorithms (Lesk and Extended Lesk), \cite{basile2014enhanced} developed an unsupervised language-independent WSD system. Instead of counting mutual words between context and sense glosses of the target word, the system uses distributional semantic space to compute the similarity between context and sense glosses. They also utilize sense frequency information from SemCor to give higher priority to most frequent senses.
    \item UKB \newline
     \cite{agirre-etal-2018-risk, agirre2009personalizing} is a graph-based WSD system which applies PageRank algorithm over a semantic graph, constructed by WordNet. In the constructed graph, the nodes and edges are WordNet synsets and relations, respectively. The algorithm assigns a PageRank value to the nodes and chose the node with the highest value as the best meaning of each target word. Two main variants of the algorithm are ppr and pprw2w. The first approach performs random walk on a graph personalized on the word context and disambiguates all the context words in one go. However, the latter performs the disambiguation process for each word separately
\end{itemize}
Our comparison also includes best supervised systems, reported in \cite{rouhizadeh-etal-2021-persian}, which utilized Persian SemCor as training set. \cite{rouhizadeh-etal-2021-persian} employed four machine learning algorithms, i.e. Support Vector Machine (SVM) (Cortes and Vapnik, 1995), K-Nearest Neighbor (KNN) (Altman, 1992), Decision Tree (DT) (Black, 1988), and Multilayer Perceptron (MLP) (McCulloch and Pitts, 1943) to train supervised Persian WSD models on Persian SemCor. All the systems make use of word embedding models as feature vector. Following \cite{rouhizadeh-etal-2021-persian}, we consider MFS as the baseline of supervised approaches. For each target word, the approach selects the most occurring sense in Persian SemCor as the most probable one.
\subsection{Parameters Settings}\label{sec:params}
To carry out the experiments in a fair setup, we first optimized the parameters of the systems on the tuning set of SBU-WSD-Corpus.\newline Among Knowledge-based systems, the pipeline of both Extended Lesk and Basile14 WSD systems, only include one parameter to tune, i.e. context size. We used the available implementations to evaluate both systems with context sizes $3$, $5$, $10$, $20$, and the whole text. Interestingly, for both systems, the best results (reported in table \ref{table_result_111}) obtained by context size $3$\footnote{The codes of Extended Lesk and Basile14 are available at https://github.com/pippokill/lesk-wsd-dsm and https://github.com/alvations/pywsd \citep{pywsd14}, respectievly}. As mentioned in \cite{agirre-etal-2018-risk}, UKB include no parameter to tune. To evaluate the system, we used the last implementation, provided by the authors of the original paper\footnote{The code is available at https://github.com/asoroa/ukb}. For supervised systems, we report the best results reported in \cite{rouhizadeh-etal-2021-persian}. \newline
All supervised systems and also Basil14 system make use of a word embedding model to represent the target texts in a semantic space. As an unsupervised machine learning techniuqe, word embedding models (e.g, word2vec\citep{mikolov2013distributed}, Glove \citep{pennington2014glove}, BERT\citep{DBLP:journals/corr/abs-1810-04805}) make use of large collections of unlabeled data to specify similar n-dimensional vectors to the semantically similar words. In order the systems with Persian, we used  Gensim software package \citep{rehurek_lrec} to train a  300-dimensional word2vec model \citep{mikolov2013distributed} on our in-house Persian news corpora.

\subsection{Evaluation Measure} \label{eval_mes}
As mentioned in \cite{navigli2009word} the performance of WSD systems can be evaluated by four standard metrics, described in the following:
\begin{enumerate}
\item Coverage \newline 
The coverage ($C$) of a WSD system is defined as the number of sense assignments provided by the system over the number of words in the test corpus. 
\begin{equation}
        C = \frac{\#\ sense\ assignments}{ \#\ total\ instances\ of\ the\ test set}
\end{equation}

\item Precision \newline
The precision (P) is defined as the number of correctly disambiguated words over the total number of disambiguated words returned by the system. 
\begin{equation}
    P = \frac{\#\ correctly\ disambiguated\ words}
    {\#\ disambiguated\ words}
\end{equation}

\item Recall \newline
The recall (R) of a WSD system is the number of the correct answers provided by the system divided by the number of expected answers.
\begin{equation}
    R = \frac{\#\ correctly\ disambiguated\ words}{\#\ total\ instances\ of\ the\ test\ set }
\end{equation}
\item F-measure \newline
F-measure is defined as harmonic mean of P and R and is computed as follows:
\begin{equation}
    F-Score = \frac{2* P * R}{P + R}
\end{equation}
Note that F-measrue = R = P, when a system provides an answer for each word in the test set.
Following Senseval-2 guidelines, we evaluate the performance of the systems with F-measure.
\end{enumerate}

\section{Results and Analysis} \label{sec:res}
Table \ref{general_sta_table} shows the F-Measure performance of all comparison systems on the SBU-WSD-Corpus dataset. We additionally, report the performance of each system, divided by PoS tags. As it can be seen, supervised systems, trained on Persian SemCor consistently outperform knowledge-based systems across the dataset. It clearly shows the high ability of Persian SemCor on training WSD models for Persian. It is also interesting to note the performance of the MFS approach, which is considered as the baseline of supervised systems, can achieve competitive results with the best performing knowledge-based systems. One of the main conclusions that can be taken from the evaluation is the positive effect of word embedding models in disambiguating Persian words. 
As discussed in section\ref{sec:params}, all the supervised models and also Basile14 utilize word embedding models in their disambiguation pipelines.
% As discussed in section \ref{sec:params}, all the supervised models utilize word embedding models in the training phase. In addition, the best-performing knowledge-based system, Basile14, uses word embedding as the main element in its disambiguating pipeline.
We provide a detailed analysis of the performance of Basile14, as the best performing knowledge-based model, to clearly show the effect of the word embedding model in its default pipeline.\newline
 As mentioned in section \ref{sec:eval_sys}, the disambiguation pipeline of Extended Lesk and Basile14 systems are highly similar. A comparison between the results obtained by these systems indicates that the use of word embedding can have a significant impact on the performance of the system. As it can be seen in table \ref{table_result_111}, the performance of Basile14 improved by a large margin (12 percent), compared to the Extended Lesk.  As discussed in section \ref{sec:eval_sys}, the pipeline of Basile14 includes two key components: (1) Word Embedding model and (2) gloss definitions of the sense inventory, both are available for Persian. Existing mutual words in the gloss of different senses which result in similarity in their semantic vectors can be mentioned as the most important bottleneck of the system. To deal with this, the system expands the gloss of each sense by including the glosses of the semantically related concepts (i.e. the concepts which have a direct relation to the synset) (see more details on section \ref{sec:eval_sys}). \newline
 We also reported the performance of the systems, divided by PoS tags. As it can be seen from table \ref{table_result_111} the performance of most systems on disambiguating nouns is lower than other PoS tags. It can be explained by the ambiguity level of different PoS tags, shown in table \ref{general_sta_table}. As shown in table \ref{general_sta_table}, the average ambiguity of the present nouns in SBU-WSD-Corpus is 4.0 which is greater than all the other PoS tags. Additionally, in Figure 2, we showed that more than 25 percent of nouns have more than 6 senses, indicating the difficulty of noun disambiguation in the developed data set. On the other hand, adjectives and adverbs seem easier to disambiguate, as their ambiguity level is 1.6 and 1.2 respectively

\section{Conclusion}
In this paper, we presented a standard evaluation corpus for Persian All-words WSD.  The corpus contains $19$ Persian documents, manually tokenized, lemmatized, PoS-tagged, and senses-tagged. It contains $5892$ words of running text and covers different domains including Economical, Sports, etc.\newline Additionally, we applied several supervised and knowledge-based WSD systems on the corpus. The results show that the supervised systems can outperform the knowledge-based alternatives. We evaluated several benchmark All-words WSD models on SBU-WSD-Corpus, providing baselines for future improvements on the Persian All-words WSD task. In addition, to encourage future research on Persian All-words WSD, we have made SBU-WSD-Corpus freely available. A possible extension to this work will include applying other knowledge-based WSD methods which are applicable to low-resource languages.

\bibliography{eacl2021.bib}
\bibliographystyle{acl_natbib}
\end{document}